\documentclass[lettersize,journal]{IEEEtran}
\usepackage{cite}
\usepackage{times}
\usepackage{mathptmx}
\def\allfiles{}
\usepackage{amsmath,amsfonts}
\usepackage{algorithmic}
\usepackage{algorithm}
\usepackage{array}
\usepackage{subfig}
\usepackage{textcomp}
\usepackage{stfloats}
\usepackage{url}
\usepackage{verbatim}
\usepackage{graphicx}
\usepackage{picins}

\usepackage[dvipsnames]{xcolor}
\definecolor{DeltaColor}{rgb}{0.039,0.73,0.71}
\definecolor{SetaColor}{rgb}{0.867, 0.0235, 0.376}
\definecolor{SigmaColor}{rgb}{0.98,0.45,0.0}
\definecolor{RedColor}{rgb}{0.8,0,0}
\definecolor{AlphaColor}{rgb}{0,0,0.8}
\definecolor{BetaColor}{rgb}{0.8,0,0.8}
\definecolor{GammaColor}{rgb}{0.5,0,0.7}
\definecolor{EpsilonColor}{rgb}{0.353,0.725,0.906}
\definecolor{TauColor}{rgb}{0.423,0.235,0.192}
\definecolor{WtColor}{rgb}{0.235,0.470,0.470}

\newcommand{\red}[1]{{\color{red}#1}}

\newcommand{\TODO}[1]{\textbf{\color{red}[TODO: #1]}}
\renewcommand{\red}[1]{{#1}}
\renewcommand{\TODO}[1]{}

\usepackage{CJKutf8}

\usepackage{mfirstuc}

\newcommand{\mynote}[1]{{\color{AlphaColor}\begin{CJK}{UTF8}{gbsn}#1\end{CJK}}}
\newcommand{\modify}[1]{{\color{DeltaColor}\begin{CJK*}{UTF8}{gbsn} #1 \end{CJK*}}}

\newcommand{\new}[1]{{#1}}

\usepackage{multirow}
\usepackage{bigstrut}
\usepackage{booktabs}

\def\mytitle{Dual-domain Adaptation Networks for Realistic Image Super-resolution}

\usepackage{soul}

\newcounter{problem}[section]
\newenvironment{problem}[1][]{\refstepcounter{problem}\par\medskip \noindent \textbf{Comment~\theproblem: #1} \rmfamily }{\medskip \vspace{-2pt}
}

\usepackage{ragged2e}

\newcounter{review}[section]

\AtBeginDocument{\numberwithin{problem}{review}}

\graphicspath{{sources/}}

\usepackage{xcolor,colortbl}

\definecolor{maroon}{cmyk}{0,0.87,0.68,0.32}

\begin{document}

\title{Dual-domain Adaptation Networks for Realistic Image Super-resolution}

\author{Chaowei Fang,~\IEEEmembership{Member,~IEEE}, Bolin Fu, De Cheng, Lechao Cheng, Guanbin Li,~\IEEEmembership{Member,~IEEE}
\thanks{Manuscript received September 10, 2024. This work was supported in part by the National Key R\&D Program of China under Grant 2024YFB3908503, in part by the National Natural Science Foundation of China under Grant 62376206, Grant 62322608 and Grant 62176198, and in part by the Key R\&D Program of Shaanxi Province under Grant 2024GX-YBXM135. (\textit{Corresponding author: De Cheng})}
\thanks{Chaowei Fang and Bolin Fu are with the Key Laboratory of Intelligent Perception and Image Understanding of Ministry of Education, School of Artificial Intelligence, Xidian University, Xi’an, China. (e-mail: chaoweifang@outlook.com, 1179502349@qq.com)}
\thanks{De Cheng is with the School of Telecommunications Engineering, Xidian University, Xi’an, Shaanxi 710071, China (e-mail: dcheng@xidian.edu.cn).}
\thanks{Lechao Cheng is with School of Computer Science and Information Engineering, Hefei University of Technology, Hefei 230601, China.}
\thanks{Guanbin Li is with the School of Computer Science and Engineering, Sun Yat-sen University, and is also with Guangdong Key Laboratory of Big Data Analysis and Processing.}
}

\markboth{IEEE Transactions on Multimedia,~Vol.~xx, No.~x, xxxx~xxxx}%
{Fang \MakeLowercase{\textit{et al.}}: Dual-domain Adaptation Networks}


\maketitle

\begin{abstract}
Realistic image super-resolution (SR) focuses on transforming real-world low-resolution (LR) images into high-resolution (HR) ones, handling more complex degradation patterns than synthetic SR tasks. This is critical for applications like surveillance, medical imaging, and consumer electronics. However, current methods struggle with limited real-world LR-HR data, impacting the learning of basic image features. Pre-trained SR models from large-scale synthetic datasets offer valuable prior knowledge, which can improve generalization, speed up training, and reduce the need for extensive real-world data in realistic SR tasks.
In this paper, we introduce a novel approach, \textit{Dual-domain Adaptation Networks}, which is able to efficiently adapt pre-trained image SR models from simulated to real-world datasets. To achieve this target, we first set up a spatial-domain adaptation strategy through selectively updating parameters of pre-trained models and employing the low-rank adaptation technique to adjust frozen parameters. Recognizing that image super-resolution involves recovering high-frequency components, we further integrate a frequency domain adaptation branch into the adapted model, which combines the spectral data of the input and the spatial-domain backbone's intermediate features to infer HR frequency maps, enhancing the SR result. Experimental evaluations on public realistic image SR benchmarks, including RealSR, D2CRealSR, and DRealSR, demonstrate the superiority of our proposed method over existing state-of-the-art models. Codes are available at: \red{\url{https://github.com/dummerchen/DAN}}.
\end{abstract}



\section{Introduction}
\label{sec:intro}
\begin{figure}[t]
\centering
\includegraphics[clip,width=1\linewidth]{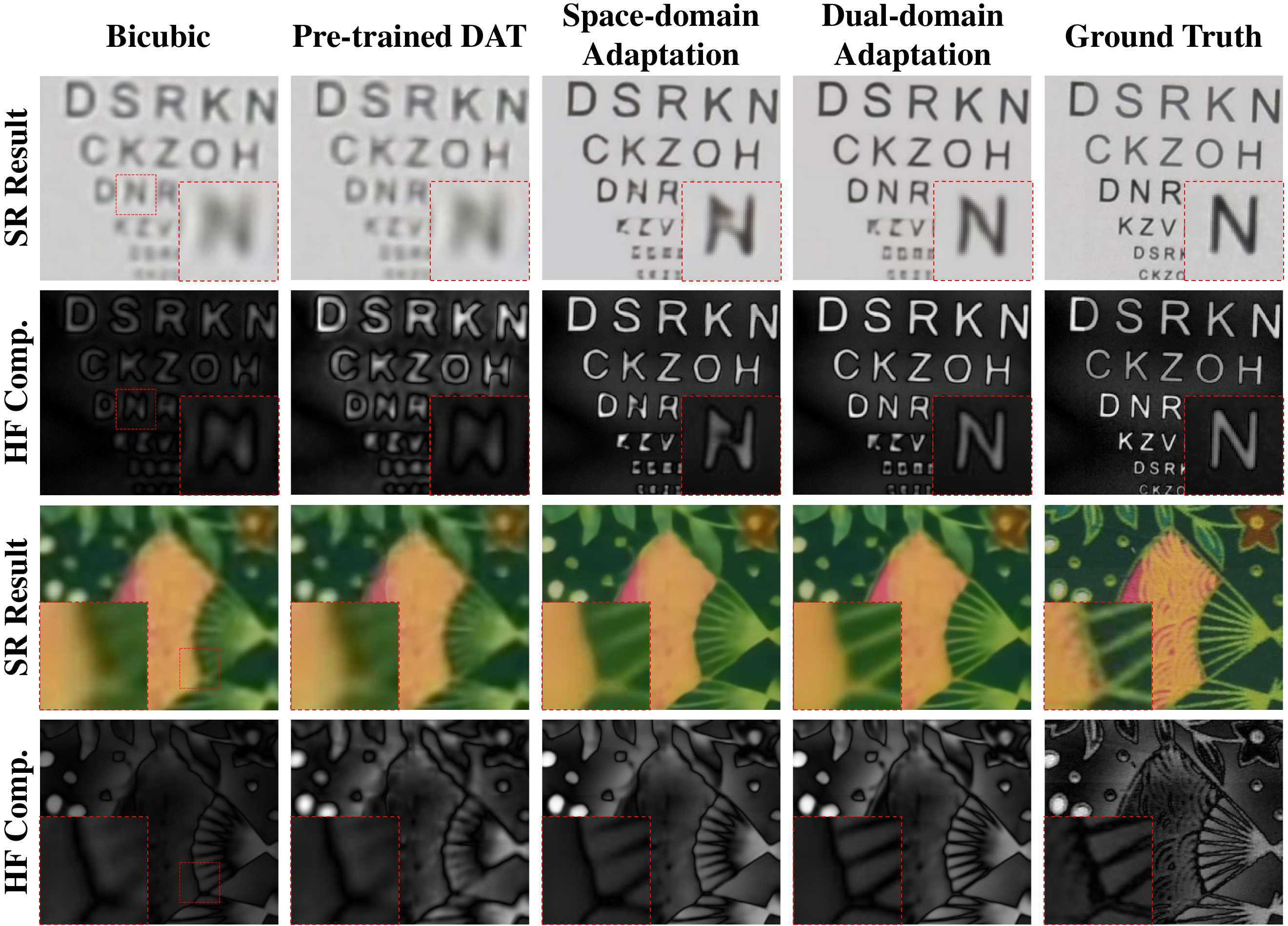}
\caption{
Our dual-domain adaptation method aims to adapt pre-trained image super-resolution (SR) models from simulated to realistic datasets. The first and third rows show SR enhancements, and the second and fourth rows highlight improved high-frequency components. The second column presents the pre-trained DAT model outcomes~\cite{chen2023dual}. The third and fourth columns demonstrate visual and structural improvements achieved with our spatial and dual-domain adaptation strategies, respectively. 
}
\label{fig:teaser}
\end{figure}




 
\IEEEPARstart{R}{ealistic} image super-resolution (SR) aims to transform low-resolution (LR) inputs from the real world to high-resolution (HR) images. 
In contrast to traditional super-resolution~\cite{dong2015image,zhang2018residual,chen2023activating}, which often deals with clean and synthetic datasets, realistic image SR focuses on handling real-world degradation patterns, making it much more applicable to actual applications. It focuses on enhancing overall image quality, recovering finer details and textures in a natural and perceptually appealing manner. This is important for improving user experiences in electronic devices such as TVs and smartphone photographs. It also has huge significance in applications such as security and surveillance, medical imaging, and forensic analysis, where improving image quality can lead to better decision-making, diagnosis, or analysis. 
Hence, this paper focuses on efficiently learning neural network-based models for addressing the realistic image SR task.

Recent methods~\cite{zhang2021designing,wang2021real} are dedicated to improving the robustness of SR models by simulating the image degradation process to generate LR images from their HR counterparts. 
However, the real image degradation process remains more complex than these simulations. An emerging solution lies in harnessing directly captured LR-HR image pairs with diverse camera configurations~\cite{chen2019camera,cai2019toward,wei2020component,li2022d2c}. The pioneering works~\cite{cai2019toward,wei2020component,li2022d2c} introduce methods focusing on leveraging adjacent information, improving the reconstruction of textural nuances, and modeling disparate high-frequency distributions.
Considering image SR models pre-trained on large-scale simulated datasets are exposed to abundant images, they are advantageous at capturing these basic features.
Intuitively, such prior knowledge of pre-trained image SR models can be used for improving the generalization ability, accelerating the training process, and relieving the burden of training data collection quality when coping with the realistic image SR task.

\new{To address the challenges of realistic image SR, we propose a novel \textit{\textbf{Dual-domain Adaptation Network}} (DAN) that efficiently adapts pre-trained SR models on simulated datasets to realistic scenarios. Our approach introduces two key innovations: a \textbf{spatial-domain adaptation strategy} and a \textbf{frequency-domain adaptation branch}, designed to synergistically enhance SR performance. Image SR models pre-trained on simulated datasets excel in extracting fundamental features, but their direct application to realistic datasets is suboptimal due to domain-specific variations. To bridge this gap, our spatial-domain adaptation strategy refines the pre-trained model by selectively freezing intermediate module parameters while updating others. This selective freezing strategy achieves an optimal balance between performance and retraining efficiency, outperforming the traditional approach of freezing only shallow modules. Additionally, we incorporate low-rank adaptation (LoRA)~\cite{hu2021lora} to fine-tune static parameters with minimal computational overhead. These techniques enable effecctive and efficient adaptation of pre-trained SR models to realistic datasets.} 
Fig.~\ref{fig:teaser} provides two examples for illustrating the effect of our proposed method in adapting a pre-trained image SR model, DAT~\cite{chen2023dual}. As shown by the third column of Fig.~\ref{fig:teaser}, the above spatial-domain adaptation strategy can significantly improve the SR results of the pre-trained DAT, while minimal extra trainable parameters are introduced.


\new{High-frequency detail restoration is critical for high-quality SR but is inadequately addressed by spatial-domain techniques alone. This may lead to memorizing pixel values without explicitly capturing intricate high-frequency textural features.} Previous researches~\cite{fuoli2021fourier,wang2023spatial,liu2023spectral} underscore the potential for image restoration based on the frequency domain, which exhibits advantages in recovering intricate high-frequency details compared to techniques based solely on the spatial domain. 
Exploiting this insight, we incorporate a frequency-domain adaptation branch into our model.
Unlike existing frequency domain methodologies~\cite{wang2023spatial,liu2023spectral}, our approach merges the Fourier transform coefficients of the input images and intermediate features produced by the spatial-domain backbone model to infer HR frequency domain images. Subsequently, transforming these images to the spatial domain yields the final prediction.
As illustrated by the fourth column of Fig.~\ref{fig:teaser}, the above frequency-domain adaptation branch helps enhance the SR results by generating more accurate and clear high-frequency components. 
Comprehensive evaluations of public realistic image SR benchmarks, including RealSR~\cite{cai2019toward}, D2CRealSR~\cite{li2022d2c}, and DRealSR~\cite{wei2020component}, confirm the superior performances of our method over existing state-of-the-art realistic image SR solutions.

Our main contributions are summarized as follows:
\begin{itemize}
\item We introduce a dual-domain adaptation networks that is able to seamlessly transfer models from simulated to real-world image SR datasets.
\item We devise a frequency-domain adaptation branch which can be flexibly integrated with existing spatial-domain backbone models, enhancing the restoration of high-frequency components.
\item Through extensive benchmark testing on public realistic image SR datasets, our method establishes a new state-of-the-art in SR performances.
\end{itemize}

\section{Related Work}
\label{sec:related}

\subsection{Realistic Image Super-resolution}
Image super-resolution, which seeks to increase the spatial resolution of images, attracts significant attention. Dong et al.~\cite{dong2015image} pioneer the introduction of the first deep neural networks (DNN) for the image SR task. With advances in DNN architectures, piles of models\cite{shi2016real,kim2016deeply,lai2017deep,lim2017enhanced,zhang2018image,fang2019self,zhu2020pnen,ma2021structure,liu2021iterative,fang2022cross,chen2023activating,choi2023n,li2023cross,chen2023dual,zhou2024ristra,liu2024cte,ran2024knlconv,liu2024degradation} are proposed with the aim of improving the performance of image SR.


To enhance the reality of image SR results, Ledig et al.~\cite{ledig2017photo} advocate for regularizing the VGG~\cite{simonyan2014very} feature distance and incorporating the generative adversarial learning loss~\cite{goodfellow2014generative} for constraining the optimization of network parameters. 
Fuoli et al.~\cite{fuoli2021fourier} present a Fourier space regularization loss to highlight the recovery of omitted frequency components. In particular, Liang et al.~\cite{liang2022details} dynamically detect visual anomalies in super-resolution images, emphasizing model learning in these regions.
Li et al.~\cite{li2022real}  focus on addressing the conflict between perceptual and pixel-reconstruction-based objectives with exclusionary masks and devise a data distillation strategy to select simulated training data having similar noise patterns with the target dataset.
Liu et al.~\cite{liu2023spectral} introduce a hybrid framework, interweaving spatial and frequency learning, employing spectral prediction uncertainty to combine the strengths of PSNR-oriented and adversarial learning-based SR models.
However, a significant portion of existing SR research is heavily based on simulated training datasets, which are typically derived from image degradation operations such as blurring and interpolation. These operations oversimplify the intricate process inherent to real LR image formation, resulting in models that can not perform well on realistic datasets.

To address the challenge of realistic image SR, Chen et al.~\cite{chen2019camera} collect real LR-HR image pairs by using various camera lenses to capture indoor postcard images. Both Cai et al.~\cite{cai2019toward} and Wei et al.~\cite{wei2020component} expand on this, establishing more extensive realistic SR datasets spanning indoor and outdoor scenes. While these datasets focus on upsampling factors of 2, 3, and 4, Li et al.~\cite{li2022d2c} introduce a $8\times$ SR dataset. The intricacies of capturing real-world images and ensuring pixel-level alignment make the creation of such datasets a laborious endeavor. Conversely, simulated image SR datasets are more straightforward to generate, and models pre-trained on them hold valuable insights that could improve realistic SR models. The efficient model adaptation from simulated to realistic datasets remains a relatively under-explored territory in image SR. 
This paper introduces a dual-domain adaptation network to bridge this gap. A spatial-domain adaptation strategy based on selective parameter updation and low-rank parameter adjustment is devised for efficiently transferring the pre-trained model. Moreover, a frequency-domain adaptation branch is incorporated for further amplifying the model's capability in high-frequency detail recovery.

\new{\subsection{Frequency-domain Representations for Image SR}
Several recent methods focus on enhancing frequency-domain representations for image SR. Fuoli et al.~\cite{fuoli2021fourier} improve high-frequency content using Fourier-space supervision but primarily focus on perceptual quality with efficient models. In contrast, our method combines spatial-domain adaptation and frequency-domain adaptation to refine both spatial and frequency features, preserving low-level details from a pre-trained model while enhancing high-frequency recovery. Wang et al.~\cite{wang2023spatial} explore the complementary nature of spatial and frequency domains in a two-branch network, but our approach goes further by adapting pre-trained models with selective parameter freezing in the spatial domain, enabling better generalization on realistic datasets. Liu et al.~\cite{liu2023spectral} focus on uncertainty estimation in the frequency domain, whereas our method directly improves high-frequency component recovery through Fourier domain adaptation while maintaining spatial domain performance. In summary, while existing methods focus on frequency-domain enhancements for SR, our approach integrates dual-domain adaptation to refine both spatial and frequency features. This not only improves high-frequency detail restoration but also preserves the low-level feature extraction capabilities of the pre-trained model, providing a more comprehensive solution for realistic image SR.}


\subsection{Neural Network Adaptation}
Addressing domain-specific challenges, such as data imbalance~\cite{sun2023unbiased,johnson2019survey}, few/noisy training data~\cite{wen2024dual,song2023comprehensive,10.1145/3664647.3681300}, and domain gap reduction~\cite{fang2020dynamic,wang2018deep,wang2023select,li2024uncertainty}, is widely studied in the field of image processing and understanding.
The paradigm of transferring models pre-trained for generic tasks to specialized downstream tasks is an effective and economic solution to tackling this problem\cite{guo2019spottune,li2020transfer}.
A prevalent strategy involves fine-tuning the entire pre-trained model on the specific downstream tasks to achieve desirable results.
To enhance efficiency and alleviate computational overhead during the fine-tuning phase, Yosinski et al.~\cite{yosinski2014transferable} and He et al.~\cite{he2022masked} advocate for the fine-tuning of only the terminal layers, while keeping preceding layers intact. 
Another line of research introduces adaptation modules directly into the neural network architecture. 
For instance, Houlsby et al.~\cite{houlsby2019parameter} and Chen et al.~\cite{chen2022adaptformer} recommend the inclusion of lightweight adaptation heads within intermediate layers. 
Chen et al.~\cite{chen2024denoising} optimize the model structure by pruning redundant structures and fusing multi-order graphs.
Hu et al.~\cite{hu2021lora} use a pair of down-projection and up-projection layers to adjust parameters of neural network layers without introducing computational overhead during inference. 
Zhang et al.~\cite{zhang2020side} and Xu et al.~\cite{xu2023side} suggest the incorporation of supplementary side adaptation branches alongside original backbone architectures. A distinct approach, as showcased by Lester et al.~\cite{lester2021power} and Jia et al.~\cite{jia2022visual}, modifies pre-trained models for downstream tasks by injecting auxiliary information into the input signals.
Recent studies have also tackled visual degradation and representation across diverse scenarios. For example, Huang et al.\cite{10146484} address cross-modal retrieval via a two-stage asymmetric hashing method. Cheng et al.\cite{cheng2024continual} propose a continual learning framework for all-in-one weather removal using knowledge replay, while Cheng et al.~\cite{cheng2024progressive} introduce a contrastive learning strategy with progressive negative enhancement for robust image dehazing. Together, these works highlight the importance of adaptable models for handling real-world visual variations.


Inspired by these neural network adaptation techniques, we delve into transferring image SR models from simulated to realistic datasets with minimal effort. A dual-domain adaptation framework based on efficient spatial-domain parameter adjustment and frequency-domain feature integration is devised to implement the transfer of image SR models. 

\ifx\allfiles\undefined
\makeatletter
\providecommand*{\input@path}{}
\g@addto@macro\input@path{{../}}
\makeatother
\documentclass[lettersize,journal]{IEEEtran}
\def\subfiles{}
\usepackage{amsmath,amsfonts}
\usepackage{algorithmic}
\usepackage{algorithm}
\usepackage{array}
\usepackage[caption=false,font=normalsize,labelfont=sf,textfont=sf]{subfig}
\usepackage{textcomp}
\usepackage{stfloats}
\usepackage{url}
\usepackage{verbatim}
\usepackage{graphicx}
\usepackage{cite}

\graphicspath{{../}{../sources}}

\usepackage{xcolor,colortbl}

\definecolor{maroon}{cmyk}{0,0.87,0.68,0.32}

\hyphenation{op-tical net-works semi-conduc-tor IEEE-Xplore}

\begin{document}

\title{\mytitle}
\fi

\section{Methodology}

\begin{figure*}[t]
\centering
\includegraphics[clip,width=1\linewidth]{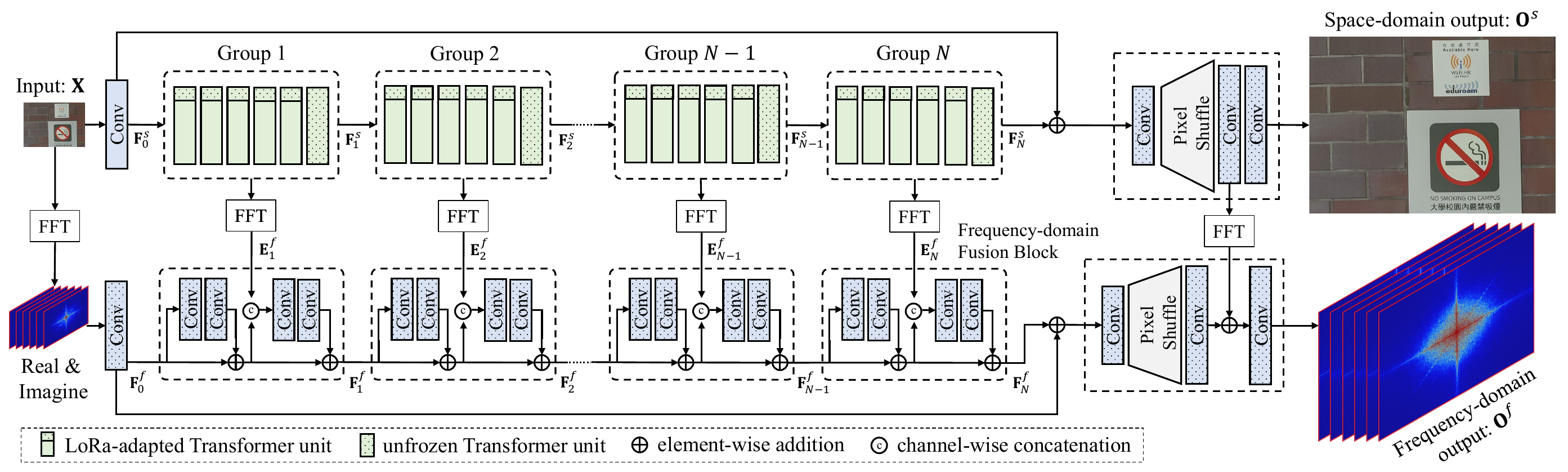}
\caption{Overview of our proposed dual-domain adaptation networks. It is built upon a pre-trained image SR model like SwinIR~\cite{liang2021swinir}, which is constituted by a head convolution, $N$ Transformer-based feature enhancement modules, and an upsampler. As shown in the upper stream, a spatial-domain adaptation strategy is introduced by unfreezing tail units of each feature enhancement module and applying low-rank adapters to adjust the remain units. The bottom stream presents the frequency-domain branch which progressively accumulates the spectral signals for enhancing the recovery of high-frequency components.
}
\label{fig:pipelinve}
\end{figure*}

This study addresses the challenge of realistic image SR. Let the input image be denoted as $\mathbf X \in \mathbb R^{ c \times h \times w}$, where $c$, $h$, and $w$ signify the number of channels, height, and width, respectively. A DNN-based image SR model is trained to predict a HR image, $\mathbf O \in \mathbb R^{ c \times \rho h \times \rho w}$, from $\mathbf X$, with $\rho$ being the upsampling ratio. The ground-truth HR image is represented as $\mathbf Y$. 

\subsection{Overview}
\new{
To overcome the limitations of existing realistic image SR approaches~\cite{cai2019toward,wei2020component,li2022d2c} that struggle with generalizing basic features and reconstructing high-frequency details, we propose a novel framework called Dual-domain Adaptation Networks (DAN). By leveraging pre-trained SR models, our method integrates both spatial-domain adaptation (SDA) and frequency-domain adaptation (FDA) to improve generalization, accelerate training, and reduce reliance on extensive paired datasets. The framework is designed to efficiently adapt pre-trained models to real-world scenarios. An overview of our framework is shown in Fig.~\ref{fig:pipelinve}.

We use SwinIR~\cite{liang2021swinir} as the pre-trained backbone model to illustrate our approach, which consists of two key components:
\begin{itemize}
    \item[1.] \textbf{Spatial-Domain Adaptation (SDA):} This branch selectively fine-tunes specific layers of the pre-trained SR model while employing low-rank adaptation (LoRa) to efficiently adjust unselected layers. This ensures effective refinement without excessive parameter updates.
    \item[2.] \textbf{Frequency-Domain Adaptation (FDA):} This branch focuses on restoring high-frequency components that are often lost in real-world low-resolution images, complementing the spatial-domain adaptations by explicitly addressing the high-frequency restoration challenges.
\end{itemize}
By combining these two strategies, the proposed framework preserves the pre-trained model's ability to capture fundamental features like edges and textures while enhancing its performance on complex, real-world degradations. This dual-domain approach delivers high-quality SR results with significantly reduced trainable network parameters.
}

\subsection{Pre-trained Backbone Model}
Models pre-trained on simulated datasets capture valuable prior knowledge for extracting basic image features necessary for super-resolving LR images.
Leveraging this prior knowledge during training on realistic datasets accelerates the process, mitigates overfitting especially with limited realistic samples, and reduces computational costs by allowing certain parameters to remain fixed.

For illustration, we use SwinIR~\cite{liang2021swinir} as our backbone model. A convolutional layer first computes a preliminary feature map, $\mathbf{F}_0^s \in \mathbb{R}^{d \times h \times w}$, from the input $\mathbf{X}$, where $d$ is the number of feature channels. This map is enhanced by $N$ groups of Transformer units, each containing $M$ units. We denote the intermediate feature maps produced by the $n$-th group of Transformer units as $\mathbf{F}_n^s \in \mathbb{R}^{d \times h \times w}$. Each Transformer unit partitions the input into local blocks. Then, it applies linear layers to compute query, key, and value variables, which are subsequently used to compute cross-pixel correlations to aggregate context information for feature enhancement. The final enhanced feature map is computed as $\mathbf{F}_{N+1}^s = \mathbf{F}_0^s + \mathbf{F}_N^s$, and an upsampler predicts the HR output $\mathbf{O}^s$ in the spatial domain.

While effective for simulated datasets, such models may underperform on realistic datasets due to differences in degradation patterns. To bridge this gap, we propose a dual-domain adaptation network (Fig.~\ref{fig:pipelinve}) that efficiently adapts the pre-trained model to realistic datasets with both spatial-domain and frequency-domain adaptation techniques.

\begin{figure}[t]
\centering
\includegraphics[clip,width=1.0\linewidth]{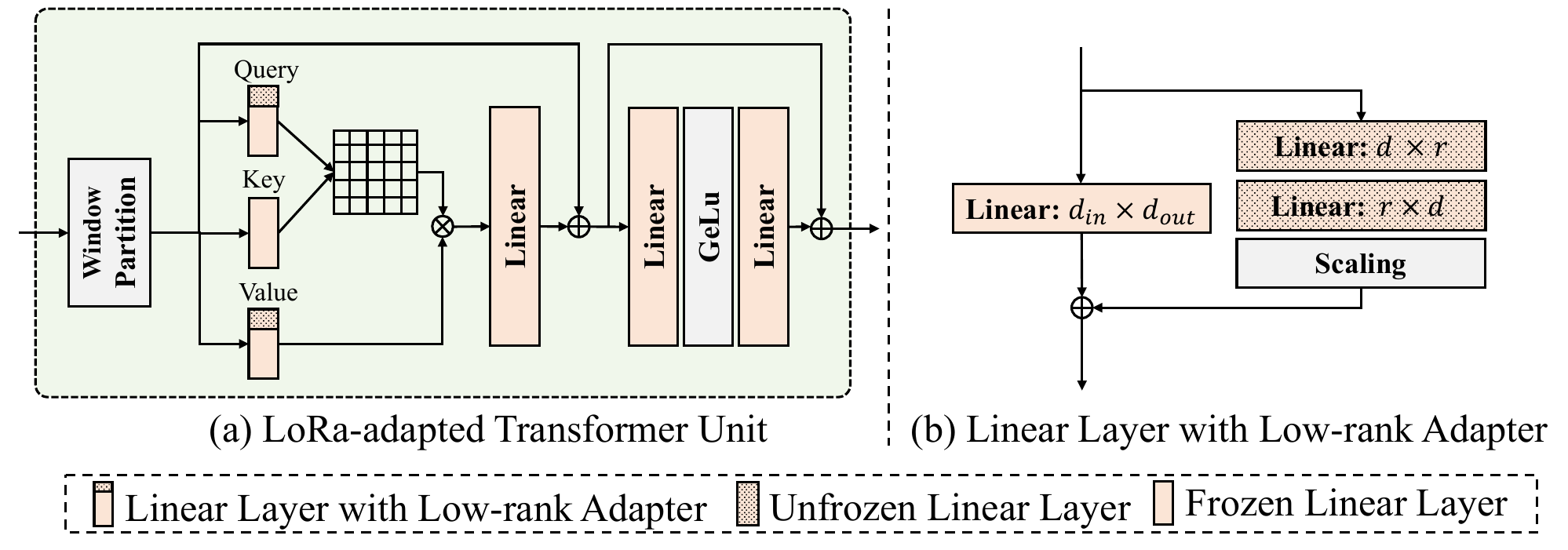}
\caption{\new{(a) Within the adapted Transformer unit, low-rank adapters are incorporated to modify the parameters of the linear layers for generating the query and value variables. The workflow of the linear layer with low-rank adapter is illustrated in (b). The output of frozen vanilla linear layer in the left branch is adapted with residuals generated by a pair of down-projection and up-projection linear layers together with a scaling layer in the right branch.}
}
\label{fig:transformer}
\end{figure}

\subsection{Dual-domain Adaptation Networks}

\subsubsection{Spatial-Domain Adaptation.}
Pre-trained models often encode general, low-level features such as edges, textures, and basic patterns in their initial layers. These foundational features are transferable across tasks and domains, making them essential for maintaining performance consistency. Freezing these layers during fine-tuning preserves this critical knowledge, preventing it from being overwritten and ensuring stable performance.
Guided by this understanding, we freeze the parameters of the convolutional head and the first $M^{sta} \in \{0,1,2,\cdots,M\}$ units within each Transformer group. Conversely, the parameters of the remaining $M^{dyn}$ units, where $M^{dyn}=M-M^{sta}$, are updated during training. This selective parameter freezing strategy enables the model to retain its ability to extract low-level features while adapting to task-specific nuances. 
Unlike completely freezing entire Transformer groups, which may introduce bottlenecks by constraining the propagation of domain-adaptive features, our devised strategy allows dynamic units to effectively process and integrate new information. This strategy mitigates overfitting and enhances the model's adaptability, ensuring improved performance on realistic datasets.

Inspired by Hu et al.~\cite{hu2021lora}, we integrate low-rank adapters to enhance the adaptability of frozen Transformer units. As delineated in Fig.~\ref{fig:transformer}, a pair of down-projection and up-projection linear layers are attached to each layer used for calculating query and value variables through the additive operation. We define $r$ as the rank value of the adapter, with $r$ being substantially smaller than $d$. \red{By integrating these adapters, we built a robust spatial domain backbone model for addressing the realistic image SR task.} Importantly, this configuration enables updates to only a restricted subset of network parameters and does not introduce additional computation demands during the inference phase.

\begin{figure}[t]
\centering
\includegraphics[clip,width=0.75\linewidth]{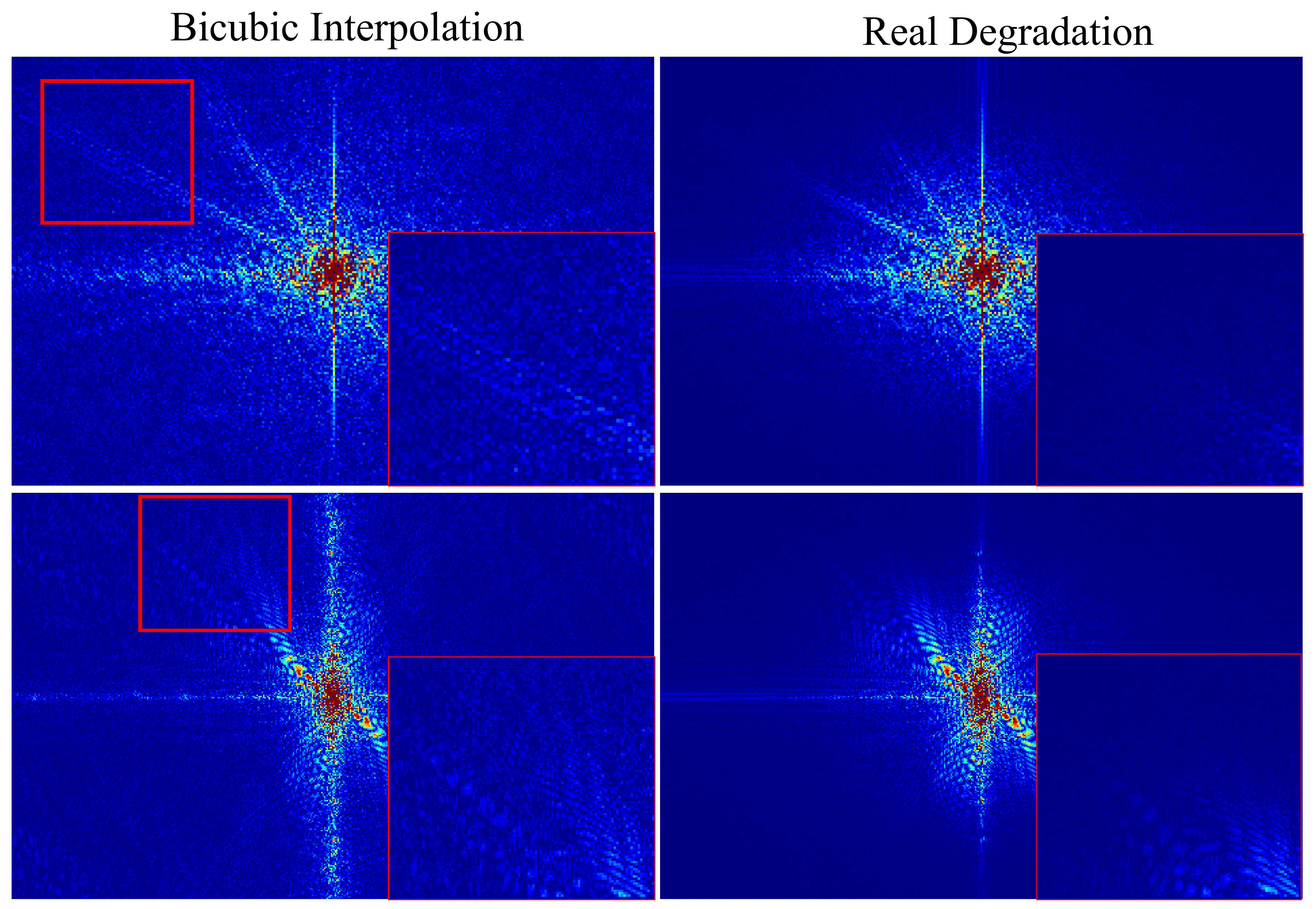}
\caption{ Visualization of frequency signals of LR images simulated by bicubic interpolation (left) and realistic LR images (right). } \label{fig:vis-freq}
\end{figure}

\subsubsection{Frequency-Domain Adaptation Branch.}
The accurate restoration of high-frequency components is pivotal in the SR image reconstruction paradigm. 
We also observe that realistic LR images often lack the high-frequency signals found in those LR images simulated via bicubic interpolation. This difference is illustrated in Fig.~\ref{fig:vis-freq}, where each row shows the frequency amplitude maps of an LR image bicubicly interpolated from a HR image in the RealSR dataset~\cite{cai2019toward} (left) and the corresponding real LR image (right). 
To enhance the capacity of the adapted SR model in restoring high-frequency components, we incorporate an auxiliary adaptation branch in the frequency domain, as depicted in the lower section of Fig.~\ref{fig:pipelinve}.

Let the real and imaginary components of the frequency domain for $\mathbf X$ be represented by $\mathbf R$ and $\mathbf I$, respectively. 
These components can be calculated utilizing the Fast Fourier Transformation (FFT) algorithm. 
A convolutional layer is used to extract the initial frequency domain feature $\mathbf F^f_0 \in \mathbb R^{d^f \times h \times w}$ from the concatenation of $\mathbf R$ and $\mathbf I$.

Subsequently, we employ $N$ fusion blocks to cumulatively combine the spectral data inherent in the intermediate features $\{\mathbf F_n^s\}_{n=1}^N$ of the spatial-domain backbone model. 
By FFT, we generate the spectral data $\mathbf E^{f}_n$, which is the concatenation of the real and imaginary components of $\mathbf F^{s}_n$. 
Each fusion block comprises two residual components: the initial component refines the frequency-domain feature map from the previous stage, namely $\mathbf F^f_{n-1}$, while the latter merges $\mathbf E^f_n$ into the refined feature map. 
This can be mathematically represented as:
\begin{equation}
\tilde{\mathbf F}^f_{n} = \mathbf F^f_{n-1} + \mathcal F_{res}( \mathbf F^f_{n-1} ), \;\; \mathbf F^f_{n} = \tilde{\mathbf F}^f_{n} + \mathcal F_{res}( [\tilde{\mathbf F}^f_{n}, \mathbf E^{f}_n] ),
\end{equation}
where $\mathcal F_{res}(\cdot)$ signifies the forward function of the residual pathway, comprised of dual convolution layers.

As illustrated in the lower right quadrant of Fig.~\ref{fig:pipelinve}, an upsampling mechanism is adopted to increase the resolution of the frequency domain feature map. The feature map produced by the penultimate convolution layer of the space-domain backbone undergoes an FFT transformation and is subsequently fused into the frequency domain's upsampler using the additive operation. The resulting high-resolution spectral maps are denoted as $\mathbf O^f$, which yield the terminal super-resolution output $\mathbf O$ upon transformation into the spatial domain.

The combination between spatial-domain and frequency-domain adaptation plays a crucial role in enhancing the performance of image SR models. The spatial-domain adaptation (SDA) contributes to transferring the low-level feature extraction capabilities of the pre-trained model into realistic scenarios with the selective parameter adjustment strategy and low-rank adapters. Meanwhile, frequency-domain adaptation (FDA) directly addresses the loss of high-frequency components, which are essential for reconstructing fine details in realistic images. By combining these two strategies, SDA focuses on refining the spatial-domain representations of basic features, while FDA enhances the restoration of high-frequency information that may be ignored during training on realistic datasets.

\subsection{Objective Function}
The cost function utilized for parameter refinement comprises two principal facets: (i) the deviation between the output of the spatial domain backbone model, i.e., $\mathbf O^s$, and the ground-truth HR image $\mathbf Y$; and (ii) the discrepancy between the output of the frequency domain adaptation branch, i.e., $\mathbf O^f$, and the Fourier coefficients of $\mathbf Y$.

Both of these deviation terms are quantified using the L1 norm. Consequently, the composite cost function can be articulated as:
\begin{equation}
L = ||\mathbf O^s - \mathbf Y||_1 + \lambda|| \mathbf O^f - \mathrm{FFT}(\mathbf Y) ||_1,
\end{equation}
where $\lambda$ ($=10$) is a constant.
Throughout the training phase, the modifiable parameters within the spatial-domain backbone model, along with entire parameters within the frequency-domain adaptation branch, undergo optimization.


\begin{table*}[t]
 \fontsize{8}{6}\selectfont
  \centering
  \caption{Comparison with existing methods on RealSR, D2CRealSR, and DRealSR datasets.
  } 
    \setlength{\tabcolsep}{1.8mm}{ \begin{tabular}{l|cc|cc|cc|cc|cc|cc|cc}
    \toprule[1pt]
    \multirow{3}[6]{*}{Method} &       \multicolumn{6}{c|}{RealSR }   & \multicolumn{2}{c|}{D2CRealSR}  &       \multicolumn{6}{c}{DRealSR }                     \bigstrut\\ \cline{2-15}          
    & \multicolumn{2}{c|}{4$\times$}  & \multicolumn{2}{c|}{3$\times$} & \multicolumn{2}{c|}{2$\times$}  & \multicolumn{2}{c|}{8$\times$}  & \multicolumn{2}{c|}{4$\times$}  & \multicolumn{2}{c|}{3$\times$} & \multicolumn{2}{c}{2$\times$} \bigstrut\\ \cmidrule{2-15}          
    &  PSNR  & SSIM & PSNR  & SSIM & PSNR  & SSIM & PSNR  & SSIM & PSNR  & SSIM & PSNR  & SSIM & PSNR  & SSIM \bigstrut\\
    \midrule
    Bicubic 
    &  27.24  & 0.764  & 28.61  & 0.810    & 31.67  & 0.887   & 27.74  & 0.822  & 30.56  &0.822    & 31.50  & 0.835   &       32.67  & 0.877     \bigstrut\\
    \midrule
    \rowcolor{gray!15}
    SRResNet\cite{ledig2017photo} 
    & 28.99  & 0.825  & 30.65  & 0.862  & 33.17  & 0.918  & 30.01  & 0.864   &31.63  & 0.847   &31.16  & 0.859   & 33.56 &  0.900 \bigstrut\\
    \midrule
    EDSR\cite{lim2017enhanced} 
    & 29.09  & 0.827  & 30.86  & 0.867 & 33.88  & 0.920 & 30.23  & 0.868  & 32.03 & 0.855 & 32.93 & 0.876  & 34.24  & 0.908   \bigstrut\\
    \midrule
    \rowcolor{gray!15}
    RCAN\cite{zhang2018image} 
    & 29.21  & 0.824 & 30.90  & 0.864  & 33.83  & 0.923 & 30.26  & 0.868 & 31.85 & 0.857  & 33.03 & 0.876  & 34.34 &  0.908    \bigstrut\\
    \midrule
    LP-KPN\cite{cai2019toward} 
    & 29.05  & 0.834  & 30.60  & 0.865  & 33.49  & 0.917  & -     & -   & 31.58 & -  & 32.64 &  -   & 33.88 &  -    \bigstrut\\
    \midrule
    \rowcolor{gray!15}
    ESRGAN\cite{wang2018esrgan} 
    & 29.15  & 0.826  & 30.72  & 0.866  & 33.80  & 0.922  & 30.06  & 0.865 & 31.92 & 0.857 & 32.39  & 0.873  & 33.89  & 0.906\bigstrut\\
    \midrule
    CDC\cite{wei2020component} 
    & 29.24  & 0.827  & 30.99  & 0.869   & 33.96  & 0.925  & 30.02  & 0.841  & 32.42 & 0.861  & 33.06 &  0.876   &  34.45 &  0.910  \bigstrut\\
    \midrule
    \rowcolor{gray!15}
    D2C-SR\cite{li2022d2c} 
    & 29.67  & 0.830  & 31.28  & 0.870   & 34.30  & 0.925  & 30.47 & 0.869  & 31.79 & 0.852 & 32.69 & 0.865 & 34.07 & 0.904 \bigstrut\\
   \midrule\midrule
    SwinIR-PreT\cite{liang2021swinir} 
    & 27.64 & 0.780 & 28.98 & 0.821 & 32.08 & 0.895 & 29.14 & 0.854  & 30.61 & 0.821 & 31.61 & 0.837 & 32.82 & 0.880  \bigstrut\\ \midrule
    \rowcolor{gray!15}
    SwinIR-ReT 
    & 29.26  & 0.829  & 30.83  & 0.865 & 34.06  & 0.926 & 30.16  & 0.873  & 31.86  & 0.850 & 32.95 & 0.869 & 34.28 & 0.906 \bigstrut\\
    \midrule
    SwinIR-FT 
    & 29.46  & 0.833  & 31.14  & 0.871 & 34.16  & 0.926  & 30.24  & 0.875 & 32.23 & 0.856 & 33.10 & 0.872 & 34.55 & 0.909  \bigstrut\\
    \midrule
    \rowcolor{gray!15}
    SwinIR-DAN-P (Ours) 
    & 29.65 & 0.835 & 31.20 & 0.872 & 34.30 &0.927 & 30.30 & 0.876 & 32.25 & 0.855 & 33.18 & 0.874  & 34.54 & 0.910   \bigstrut\\
    \midrule
    SwinIR-DAN-F (Ours) 
    & 29.70  & 0.837  & 31.31  & 0.874 & 34.40  & 0.928 & 30.36  & 0.876 & 32.45 & 0.859 & 33.34 & 0.877 & 34.75 & 0.913  \bigstrut\\
    \midrule\midrule
    \rowcolor{gray!15}
    DAT-PreT\cite{chen2023dual} 
    & 27.64 & 0.780  & 28.98 & 0.820 & 32.08 & 0.895  & - & -  & 30.61 & 0.821 & 31.60 & 0.837 & 32.82 & 0.880\bigstrut\\ \midrule
    DAT-ReT 
    & 29.51 & 0.831  & 31.10  & 0.869  & 34.19  & 0.926 & 30.04 & 0.873 & 31.90 & 0.850 & 32.91 & 0.868 & 34.28 & 0.907   \bigstrut\\
    \midrule
    \rowcolor{gray!15}
    DAT-FT 
    & 29.71  & 0.839 & 31.36 & 0.876 & 34.41  & 0.929 & 30.26  & 0.877  & 32.32 & 0.860 & 33.18 & 0.874 & 34.62 & 0.912 \bigstrut\\
    \midrule
    DAT-DAN-P (Ours) 
    & 29.95 & 0.841 &  31.59 & 0.878 & 34.58 & 0.930 & 30.49 & 0.876 & 32.36 & 0.857 & 33.20 & 0.874 & 34.63 & 0.912         \bigstrut\\
    \midrule
    \rowcolor{gray!15}
    DAT-DAN-F (Ours)  
    & \textbf{30.00} & \textbf{0.843} & \textbf{31.68} & \textbf{0.880} & \textbf{34.73} & \textbf{0.932} & \textbf{30.51}  & \textbf{0.878}  &  \textbf{32.58} & \textbf{0.862} & \textbf{33.42} & \textbf{0.880} & \textbf{34.87} & \textbf{0.915}  \bigstrut\\
    \bottomrule[1pt]
    \end{tabular}%
    }
  \label{tab:cmp1}%
\end{table*}

\begin{figure}[t]
\centering
\includegraphics[clip,width=1\linewidth, trim={0 0 320 0}]{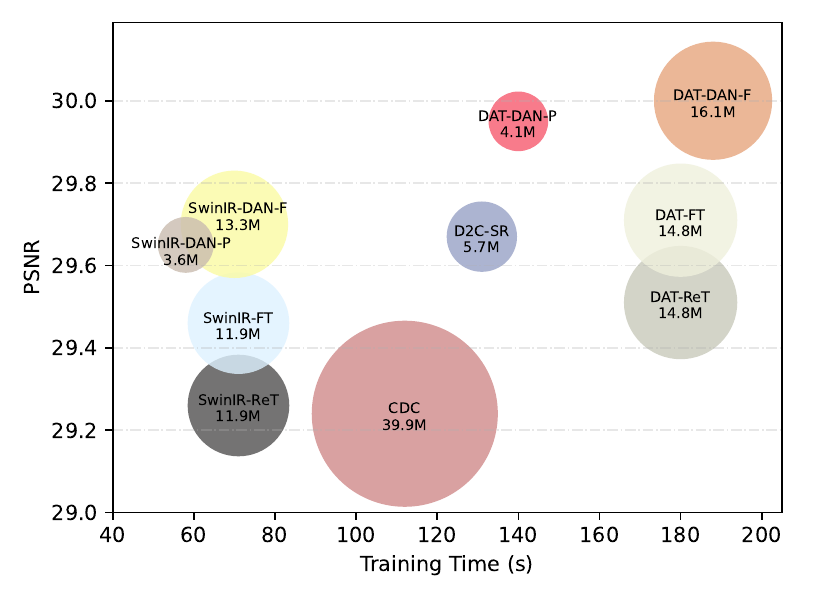}
\caption{Scatter plots of PSNR values, training time per epoch, and trainable parameter amount of different methods. Larger points indicate more trainable parameters. }
\label{fig:time-cost}
\end{figure}

\begin{figure*}[t]
\centering
\includegraphics[clip,width=1\linewidth]{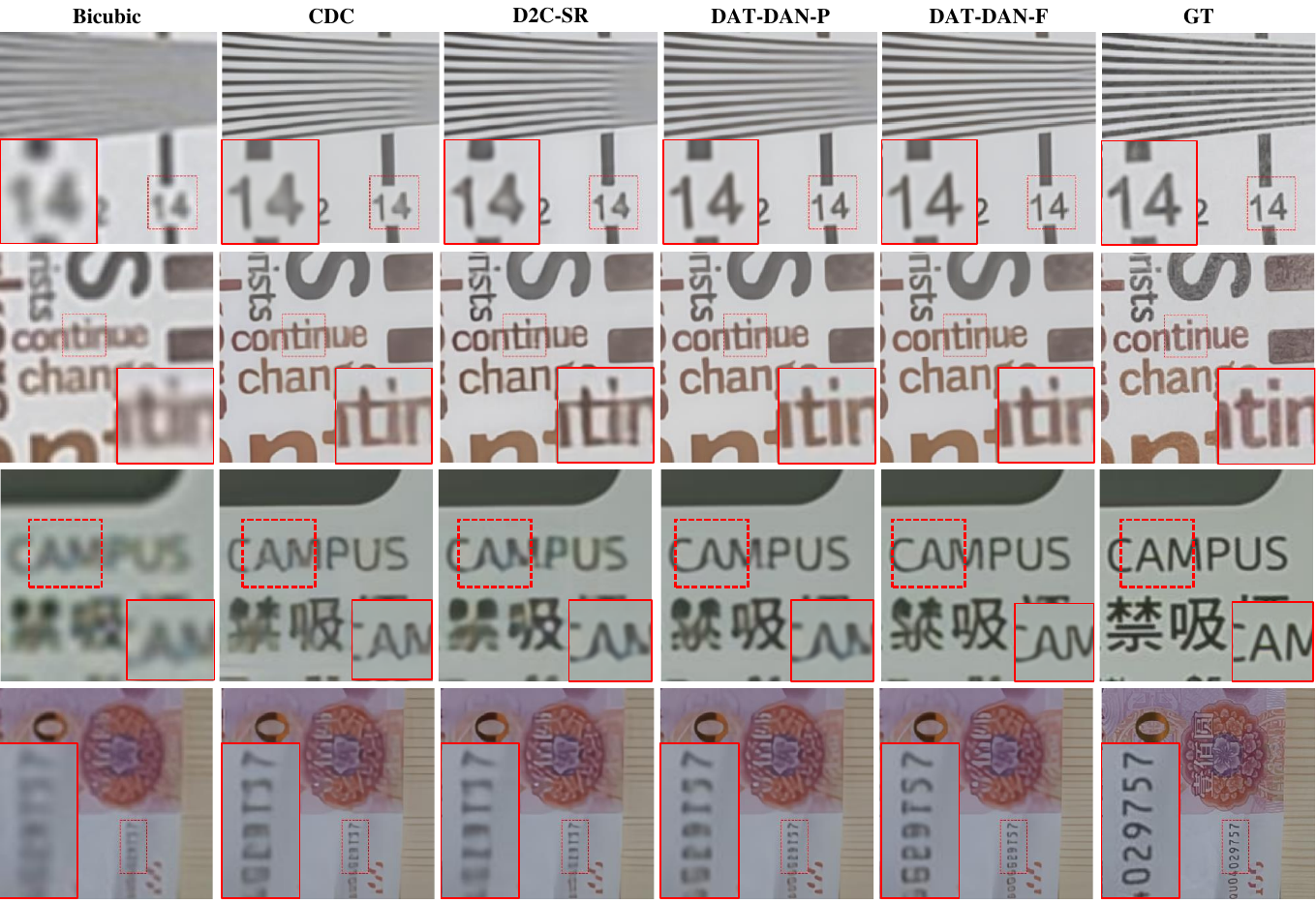}
\caption{A qualitative comparison among different models on four $4\times$ SR examples from the RealSR dataset. From left to right: bicubic interpolation, CDC~\cite{wei2020component}, D2C-SR~\cite{li2022d2c}, DAT-DAN-P, DAT-DAN-F, and the ground-truth (GT) image. 
}
\vspace{-5mm}
\label{fig:cmp-q}
\end{figure*}

\begin{figure*}[htbp]
    \centering
    \includegraphics[clip,width=1\linewidth]{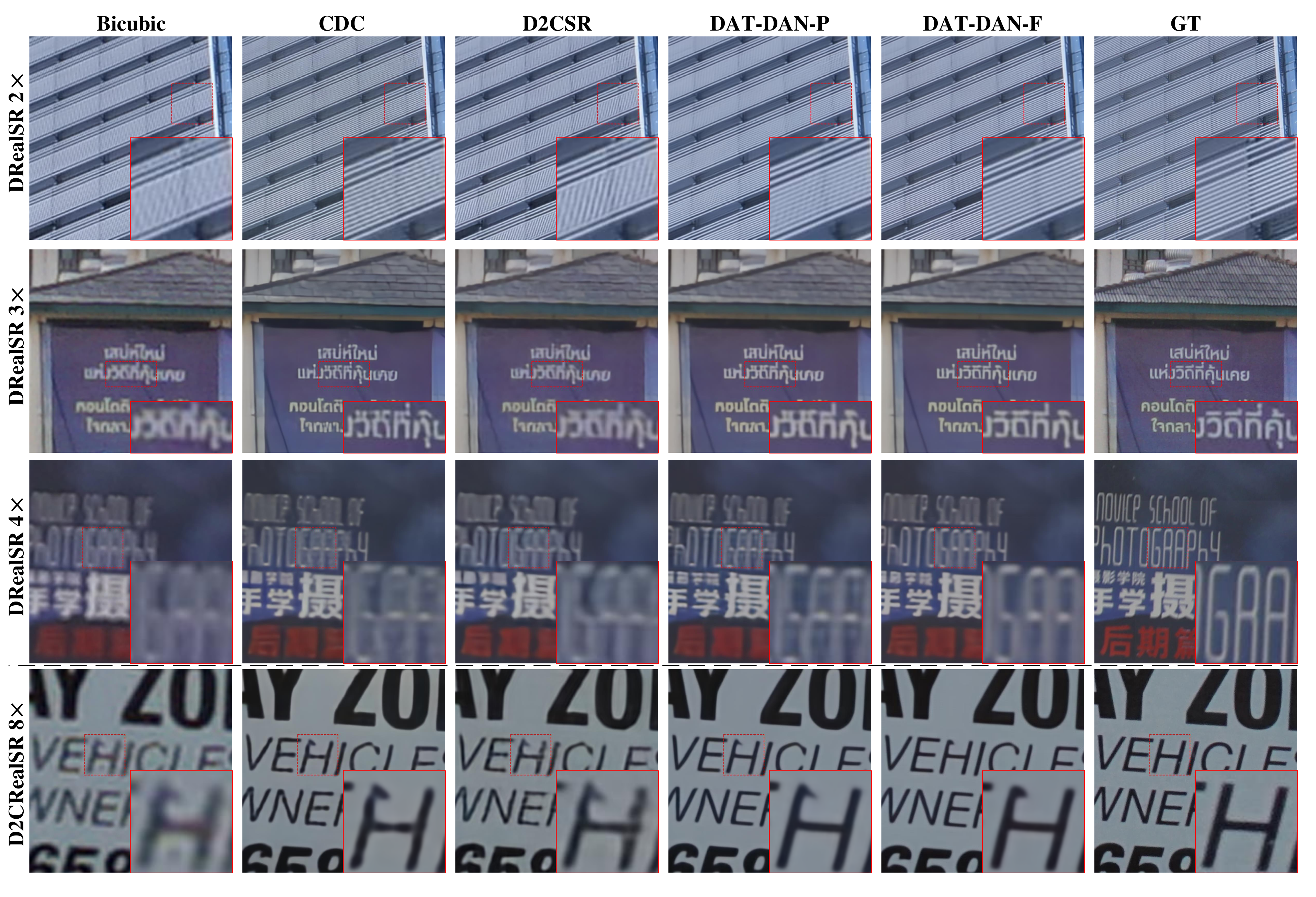}
    \caption{Qualitative comparison among different models on $2\times,3\times,4\times$ SR setting of the DRealSR dataset~\cite{wei2020component} and $8\times$ SR setting of the D2CRealSR\cite{li2022d2c}. From left to right: bicubic interpolation, CDC~\cite{wei2020component}, D2C-SR~\cite{li2022d2c}, DAT-DAN-P, DAT-DAN-F, and the ground-truth (GT) image. 
    }
    \label{fig:cmp_compress}
\end{figure*}

\section{Experiments}
\subsection{Datasets}
Three realistic SR datasets are used to evaluate the performance of SR methods: 
\begin{itemize}
    \item \textbf{RealSR}~\cite{cai2019toward} contains LR and HR image pairs captured from 559 scenes using two DSLR cameras including Canon 5D3 and Nikon D810. Every scene has an HR image and its three LR counterparts having 1/2, 1/3, and 1/4 resolutions, respectively. 459 and 100 image pairs are used for training and testing, respectively.
    \item \textbf{D2CRealSR}~\cite{li2022d2c} contains 115 pairs of LR and HR images for $8\times$ image SR. They are split into 100 and 15 for training and testing, respectively.
    \item \textbf{DRealSR}~\cite{wei2020component} contains 884, 783, and 840 training image pairs, and 83, 84, and 93 testing image pairs for $2\times$, $3\times$, and $4\times$ SR, respectively. 
\end{itemize}

\subsection{Implementation Details}
In this study, we employ PyTorch~\cite{imambi2021pytorch} for the implementation of our proposed methodology. Our experimental framework involves the integration of dual-domain adaptation networks with two distinct backbone models: SwinIR~\cite{liang2021swinir} pre-trained on the DIV2K dataset, and DAT~\cite{chen2023dual} pre-trained on a composite dataset combining DIV2K and Flickr2K.
These backbone models are architecturally composed of six feature enhancement groups (i.e., $N = 6$), with each group comprising six Transformer units (i.e., $M = 6$). The intermediate feature maps within these models have 180 channels, i.e., $d=180$.
In our default configuration, we opt to freeze the parameters of the convolution head and \new{the first five units of each Transformer group, namely $M^{sta}=5$}. 
Regarding the low-rank adapters, we assign a value of 4 to $r$. For DAT, low-rank adapters are applied to adjust the parameters of linear layers for generating key and value variables in spatial or channel-wise self-attention modules and those for constructing the spatial-gate feed-forward network.
For the frequency-domain adaptation branch, the dimension $d^f$ is set to 64.
During training, we use the Adam optimizer~\cite{kingma2014adam} to update the parameters. 
The initial learning rate is established at $2\times10^{-4}$, which is subsequently halved every 2,000 iterations. Training images are processed to randomly crop $96\times96 $ LR patches. We set the batch size at 4 and conduct the training for 70,000 iterations. 


\subsection{Comparisons with Existing Methods}

\subsubsection{Quantitative Comparison} In our comprehensive evaluation, detailed in Table~\ref{tab:cmp1}, we perform a comparative analysis using PSNR and SSIM metrics to assess our dual-domain adaptation methodology against established image SR techniques such as SRResNet~\cite{ledig2017photo}, EDSR~\cite{lim2017enhanced}, RCAN~\cite{zhang2018image}, LP-KPN~\cite{cai2019toward}, ESRGAN~\cite{wang2018esrgan}, CDC~\cite{wei2020component}, D2C-SR~\cite{li2022d2c}, along with variants SwinIR or DAT  adapted from simulated datasets to realistic ones. 
\new{The metric values of SRResNet and EDSR are taken from~\cite{li2022d2c}.} 
This assessment highlights that pre-trained models including SwinIR-PreT and DAT-PreT, show reduced effectiveness on realistic datasets. 
Attempts to re-train SwinIR and DAT from scratch as indicated by SwinIR-ReT and DAT-ReT, respectively, exhibit a notable underperformance against D2C-SR across various SR settings on the RealSR dataset.

We try to explore full fine-tuning (FT) of all parameters on the pre-trained SwinIR and DAT, forming SwinIR-FT and DAT-FT, respectively.  These two models consistently improve performance across all SR settings. 
Despite its effectiveness, the FT process incurs computational inefficiencies due to the necessity of updating all parameters. 
We then integrate our novel dual-domain adaptation networks with the pre-trained SwinIR and DAT, leading to the development of SwinIR-DAN-P and DAT-DAN-P, respectively. 
These variants, which require around 30\% of the trainable parameters of their FT counterparts, respectively, demonstrate not only a reduction in the number of trainable parameters but also outperforming the FT method, with SwinIR-DAN-P and DAT-DAN-P surpassing SwinIR-FT and DAT-FT by 0.21dB and 0.24dB, respectively, in the $4\times$ SR setting. 
In particular, DAT-DAN-P exceeds the performance of the previously best method, D2C-SR.
Our method's potential is further underscored by unfreezing all parameters in the backbone models, resulting in SwinIR-DAN-F and DAT-DAN-F variants, which show substantially improved results.

Fig.~\ref{fig:time-cost} shows the scatter plot of PSNR values, training time per epoch, and trainable parameter amount of different methods. All metrics are evaluated under the 4$\times$ SR setting of the RealSR dataset. The training time per epoch is tested with a Nvidia GeForce RTX 3090 GPU.
It can be observed that our method variant SwinIR-DAN-P or DAT-DAN-P have much less training time and trainable parameters while achieving higher PSNR value, compared to SwinIR-FT or DAT-FT, respectively. This indicates that our devised DAN is effective in improving the training efficiency compared to full fine-tuning.
Compared to D2C-SR, SwinIR-DAN-P has significantly less training time while achieving comparable PSNR value; DAT-DAN-P has comparable training time while leading to much higher PSNR value.
It can be deduced that making use of image SR models pre-trained with simulated data can effectively improve the SR performance or training efficiency for learning realistic image SR models. 

\subsubsection{Qualitative Comparison} 
\new{
Fig.~\ref{fig:cmp-q} and Fig.~\ref{fig:cmp_compress} visualize examples from the different SR settings of RealSR, DRealSR and D2CRealSR dataset, respectively.
As illustrated by the four examples of Fig.~\ref{fig:cmp-q}, our method variants DAT-DAN-P and DAT-DAN-F are capable of generating characters with sharper edges and clearer appearances than D2C-SR and CDC.
As illustrated of Fig.~\ref{fig:cmp_compress}, our method demonstrates enhanced performance in reconstructing building surface streaks and high-frequency structural details
}

\subsection{Cross-camera Adaptation}
To validate the robustness of our method in the situation of cross-camera adaptation, we conduct experiments by dividing the RealSR dataset into two subsets according to the camera, including Canon and Nikon. One subset is used to pre-train the DAT model which is subsequently adapted to the other one.
The experimental results are presented in Table~\ref{tab:crosscamera}. 
Our proposed method DAN-P performs significantly better than retraining from scratch (ReT) and full fine-tuning (FT).
For example, under the setting of Canon$\rightarrow$Nikon, the PSNR of DAN-P is 0.42 and 0.25 higher than that of ReT and FT, respectively. These experiments further validate the generalization of our method in different real-world conditions, affirming the robustness of our DAN-P and DAN-F variants. 


\begin{table}[t]
\begin{center}
\fontsize{8}{10}\selectfont
\caption{Performance in cross-camera testings on RealSR dataset, including Canon$\rightarrow$Nikon and Nikon$\rightarrow$Canon. DAT is used as the backbone model.}\label{tab:crosscamera} 
\setlength{\tabcolsep}{2mm}{ 
\begin{tabular}{l|c|c|c|c} \toprule[1pt]
Settings     & Metrics & ReT & FT & DAN-P  \\ \midrule
\multirow{2}{*}{Canon$\rightarrow$Nikon} 
& PSNR & 28.11 & 28.28 & 28.53 \\ \cmidrule{2-5}
& SSIM & 0.794 & 0.811 & 0.814  \\ \midrule
\multirow{2}{*}{Nikon$\rightarrow$Canon} 
& PSNR & 29.23 & 29.35 & 29.60  \\ \cmidrule{2-5}
& SSIM & 0.832 & 0.839 & 0.840   \\ \bottomrule
\end{tabular}
}
\end{center}
\end{table}

\begin{figure}[t]
\centering
\includegraphics[clip,width=1\linewidth]{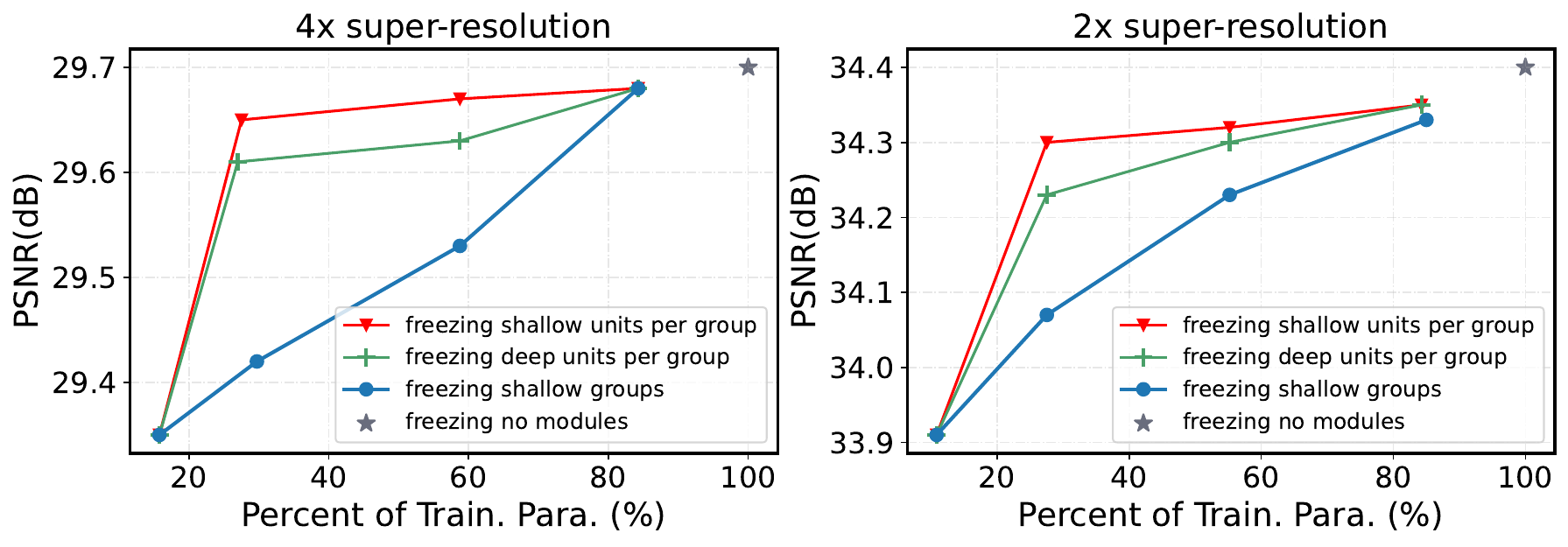}
\caption{Variation of PSNR with respect to the percent of trainable parameters.
}
\label{fig:freezing}
\end{figure}

\begin{figure}[t]
\centering
\includegraphics[clip,width=1\linewidth]{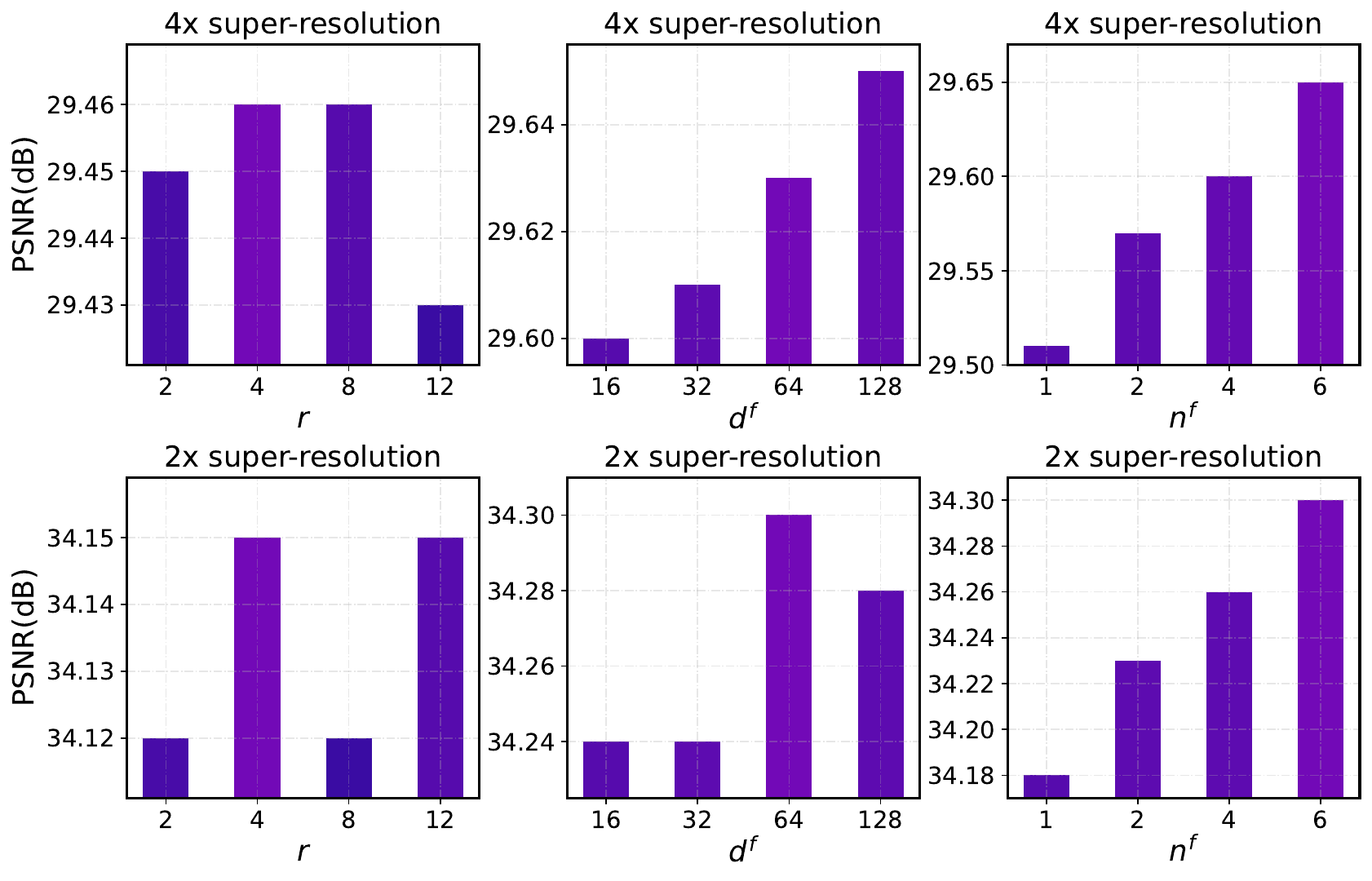}
\caption{Variation of the PSNR metric with respect to the rank value $r$, the feature dimension $d^f$ in the frequency-domain adaptation branch and the number of frequency domain adaptation stage $n^f$.  
}
\label{fig:para}
\end{figure}

\subsection{Ablation Study}
This section is dedicated to conducting comprehensive ablation experiments to validate the effectiveness of the key components in our method. We also evaluate the impact of employing diverse strategies and hyper-parameters for model adaptation. These experiments are carried out systematically on the RealSR dataset, and SwinIR is used as the space-domain backbone model.


\begin{table}[t]
  \centering
  \fontsize{8}{10}\selectfont
  \caption{Ablation study for key components in 4$\times$ and 2$\times$ settings of RealSR. 
  `SPFT', `LoRa', and `FDA' denotes selective parameter fine-tuning, low-rank adapter, and frequency-domain adaptation, respectively. 
  }
    \setlength{\tabcolsep}{2mm}{ \begin{tabular}{c|c|c|c|cc|cc}
    \toprule[1pt]
     \multirow{2}[4]{*}{No.}    &  \multirow{2}[4]{*}{SPFT} &  \multirow{2}[4]{*}{LoRa} &  \multirow{2}[4]{*}{FDA}    & \multicolumn{2}{c|}{4$\times$}   & \multicolumn{2}{c}{2$\times$}  \bigstrut\\
\cmidrule{5-8}        
    &  &  & & PSNR  & SSIM   & PSNR  & SSIM  \bigstrut\\
    \midrule
    1 &            &  &  & 29.26  & 0.829  & 34.06  & 0.926   \bigstrut\\
    \midrule
    2 & \checkmark &  &   & 29.47 & 0.831 & 34.11  & 0.925 \bigstrut\\
    \midrule
    3 & \checkmark &  \checkmark &   & 29.46 & 0.832 & 34.15 &0.925 \bigstrut\\
    \midrule
    4 & \checkmark &  &\checkmark & 29.58 & 0.834 & 34.24 &0.926   \bigstrut\\
    \midrule
    5 & \checkmark & \checkmark &\checkmark & 29.65 & 0.835  & 34.30  & 0.927  \bigstrut\\
    \bottomrule[1pt]
    \end{tabular}%
    }
  \label{tab:ablation}%
\end{table}%

\begin{table}[t]
\centering
\fontsize{7}{7}\selectfont
\caption{Performance of replacing the FDA branch in DAN-P with a spatial domain adaptation branch (SDA-P) on $4\times$ RealSR dataset.
} 
\setlength{\tabcolsep}{5mm}{ 
\begin{tabular}{c|c|c|c} \toprule[1pt]
Variants  & PSNR  & SSIM & $N^{trn}$  \\ \midrule

SDA-P     & 29.44 & 0.832 & 4.6        \\ \midrule
DAN-P     & 29.65 & 0.835 & 3.6        \\
\bottomrule[1pt]
\end{tabular} 
}
\label{tab:sda}
\end{table}

\vspace{.2em}
\noindent \textbf{$\bullet$ Efficacy of Principal Components.}
Table~\ref{tab:ablation} illustrates the implementation of various variants of our method to substantiate the effectiveness of key components including selective parameter fine-tuning (SPFT), low-rank adapters (LoRa), and frequency-domain adaptation (FDA). The pre-trained SwinIR model serves as the foundational baseline (referenced as No. 1). Upon fine-tuning selective parameters of this baseline model, encompassing the parameters of the last Transformer units in all feature enhancement groups and the upsampler (referenced as No. 2), an enhancement in the SR results is observed. For example, there is an increase of 0.21dB in the PSNR metric in the $4\times$ SR setting.

Furthermore, the incorporation of FDA, as illustrated in No. 4, results in substantial performance improvement. Specifically, there is an increase in the PSNR metric by 0.11dB and 0.13dB in the $4\times$ and $2\times$ SR settings, respectively. As depicted in No. 3 and No. 5, the use of LoRa to modulate the frozen parameters of the backbone model contributes to certain performance enhancements. We attempt to alter the FDA branch with an extra spatial-domain branch having more parameters. As shown in Table \ref{tab:sda}, this method variant indicated by (SDA-P) is unable to improve the PSNR and SSIM metrics compared to No. 3 in Table~\ref{tab:ablation}, indicating the extra spatial-domain branch is redundant in adapting the backbone model. However, our devised FDA branch can still bring improvement by enhancing the recovery of high-frequency components in the Fourier domain.

\begin{table}[t]
  \centering
  \caption{The results of replacing FFT with wavelet transform on the RealSR dataset.
  }
    \begin{tabular}{l|cc|cc}
    \toprule
    \multicolumn{1}{c|}{\multirow{2}[4]{*}{Method Variants}} & \multicolumn{2}{c|}{4$\times$} & \multicolumn{2}{c}{2$\times$} \\
\cmidrule{2-5}      & PSNR & SSIM & PSNR & SSIM \\
    \midrule
    Wavelet Transform & 29.47 & 0.832 & 34.15 & 0.925  \\ \midrule
    Fast Fourier Transform & 29.65 & 0.836 & 34.30 & 0.927  \\
    
    \bottomrule
    \end{tabular}
  \label{tab:r1c2}
\end{table}

\vspace{.2em}
\new{\noindent \textbf{$\bullet$ Comparison FFT against Wavelet Transform in Spatial-Frequency Decomposition.}
We implement a variant of our method through replacing the Fast Fourier Transform (FFT) with wavelet transform (WT). The results of this variant are presented in the third row of Table~\ref{tab:r1c2}.
Compared to FFT, WT shows a performance decline, e.g., the PSNR of WT is 0.18 and 0.15 lower than that of FFT on 4$\times$ and 2$\times$ SR settings, respectively. 
The potential reason is that FFT provides a global frequency decomposition which is useful for capturing periodic patterns, textures, and high-frequency details in a holistic manner while WT only captures local high-frequency details. This global perspective is particularly advantageous in SR tasks that require precise restoration of high-frequency components.
}

\vspace{.2em}
\new{
\noindent \textbf{$\bullet$ Choice of Parameter Freezing Strategies.}
Our analysis extends to the performance implications of employing three distinct strategies for freezing parameters of the backbone model. 
In the first strategy, we freeze the parameters of the shallowest groups completely. We label this strategy as ``freezing shallow groups'' in Fig.~\ref{fig:freezing}.
In the second strategy which is labeled as  ``freezing deep units per group'' in  Fig.~\ref{fig:freezing},  we freeze the parameters of deep Transformer units of each group while updating the parameters of other Transformer units.
The third strategy, which is the one adopted in our final method, involves freezing the first several Transformer units in each group. As indicated by the curve labeled ``freezing shallow units per group'' in Fig.~\ref{fig:freezing}, this approach yields a more favorable balance between performance and the number of trainable parameters.
}


\vspace{.2em}
\noindent \textbf{$\bullet$ Variability of Hyper-parameters.}
\new{The influence of choosing values for the rank value $r$, the feature dimension $d^f$ within the frequency-domain adaptation branch and the number of frequency-domain adaptation stages (denoted as $n^f$) is depicted in Fig.~\ref{fig:para}.}
\begin{enumerate}
\item[1)] The rank value $r$ controls the number of learnable parameters in adapters for frozen layers. Optimal SR performance is achieved by setting $r$ to 4, while larger values for $r$ do not confer additional benefits. 
\item[2)] The parameter $d^f$ determines the complexity of the FDA branch. The peak performance in $4\times$ and $2\times$ SR is attained at different values. In the $2\times$ SR context, the PSNR value plateaus beyond a $d^f$ setting of 64; in the $4\times$ SR context, the PSNR value shows a continuous increase as $d^f$ increases from 16 to 128. Balancing both performance and resource consumption considerations, we establish $d^f$ at 64 in the final version of our method.
\item[3)] Increasing the number of FDA stages $n^f$ leads to higher PSNR values, since using more FDA stages helps capture more nuanced high-frequency features.
\end{enumerate}

\begin{figure}[t]
    \centering
    \includegraphics[width=1\linewidth]{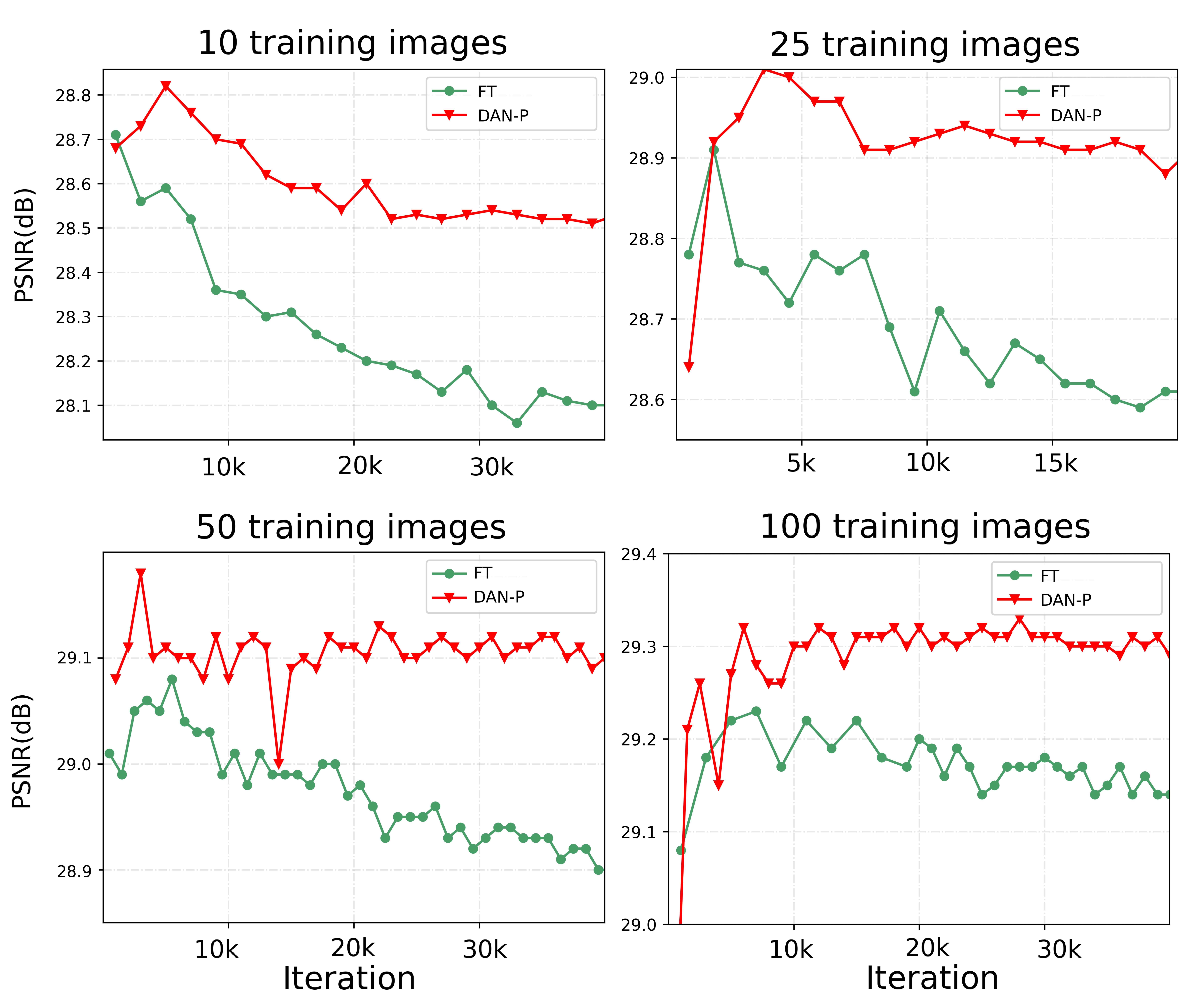}
    \caption{PSNR variation curves produced with different numbers of training images. 
    }
    \label{fig:overfit}
\end{figure}

\vspace{.2em}
\noindent \textbf{$\bullet$ Performance under Various Numbers of Training Images.}
We conduct experiments using various numbers of training images including $10$, $25$, $50$, and $100$. Fig.~\ref{fig:overfit} illustrates the PSNR variation curves of FT and DAN-P with respect to the training iteration.
The PSNR values of FT degrade substantially with increasing training iteration due to overfitting. The overfitting issue intensifies as the number of training images decreases. The PSNR values of our DAN-P remain stably as the training process advances when using 100 training images. For 50 training images, the PSNR value of DAN-P suffers a decrease at around 3,000 iterations and then saturates in subsequent training iterations.
For 25 and 10 training images, moderate overfitting issue of DAN-P can be observed. In summary, compared to FT, our DAN-P method can effectively alleviate the overfitting issue while achieving better performance.


\section{Conclusion}
This study introduces a dual-domain adaptation \new{network} for transferring image SR models from simulated to realistic datasets. 
We find that selective parameter fine-tuning and frequency domain adaptation notably improve SR performance. 
Our analysis reveals that freezing intermediate Transformer units offers a better performance-resource balance compared to freezing the shallowest modules. 
Low-rank adapters also contribute to adjusting the frozen parameters of the backbone model. 
Our devised network adaptation method significantly outperforms the full fine-tuning strategy using nearly one third of the trainable parameters.
Our method can significantly improve pre-trained backbone models, achieving new state-of-the-art performances on RealSR, D2CRealSR, and DRealSR datasets.
The limitation of this work is that the pre-trained image SR models usually have high network complexity. 
Learning light-weight image SR models with the help of the knowledge of pre-trained large models deserves future research.

\bibliographystyle{IEEEtran}
\bibliography{main.bib}

\vfill
\newpage

\par\noindent 
\parbox[t]{\linewidth}{
\noindent\parpic{\includegraphics[height=1.5in,width=1in,clip,keepaspectratio]{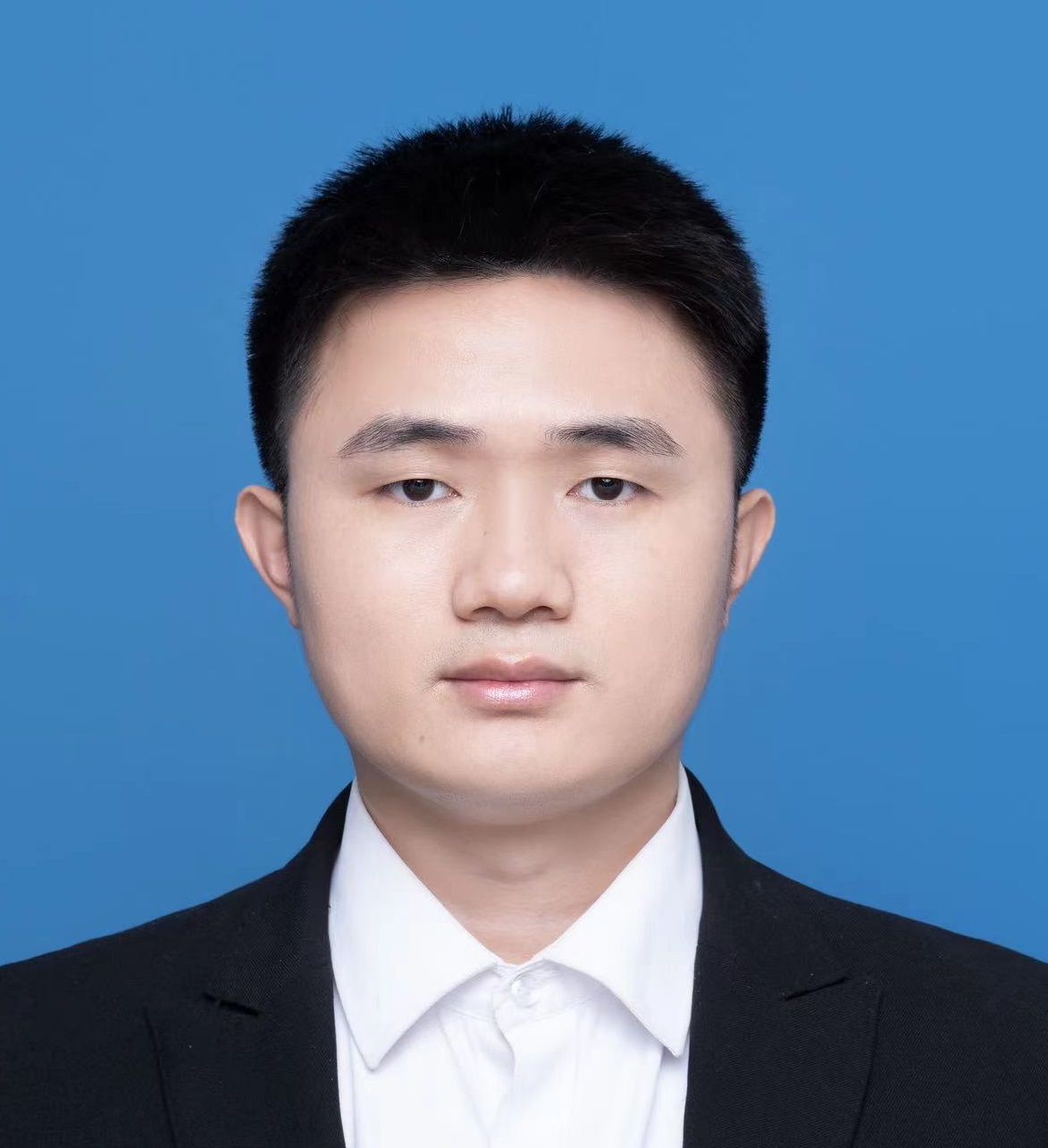}}
\noindent {\bf Chaowei Fang}\
received the Ph.D. degree from the University of Hong Kong, Hong Kong, in 2019. He is an Associate Professor at the School of Artificial Intelligence, Xidian University, Xi’an, China. He has contributed as an author or a coauthor to over 50 publications featured in prestigious journals and conferences. He served as a senior program committee member of ECAI 2024-2025. His research interests include multi-modal modeling and robust machine learning.}
\vspace{1\baselineskip}

\parbox[t]{\linewidth}{
\noindent\parpic{\includegraphics[height=1.5in,width=1in,clip,keepaspectratio]{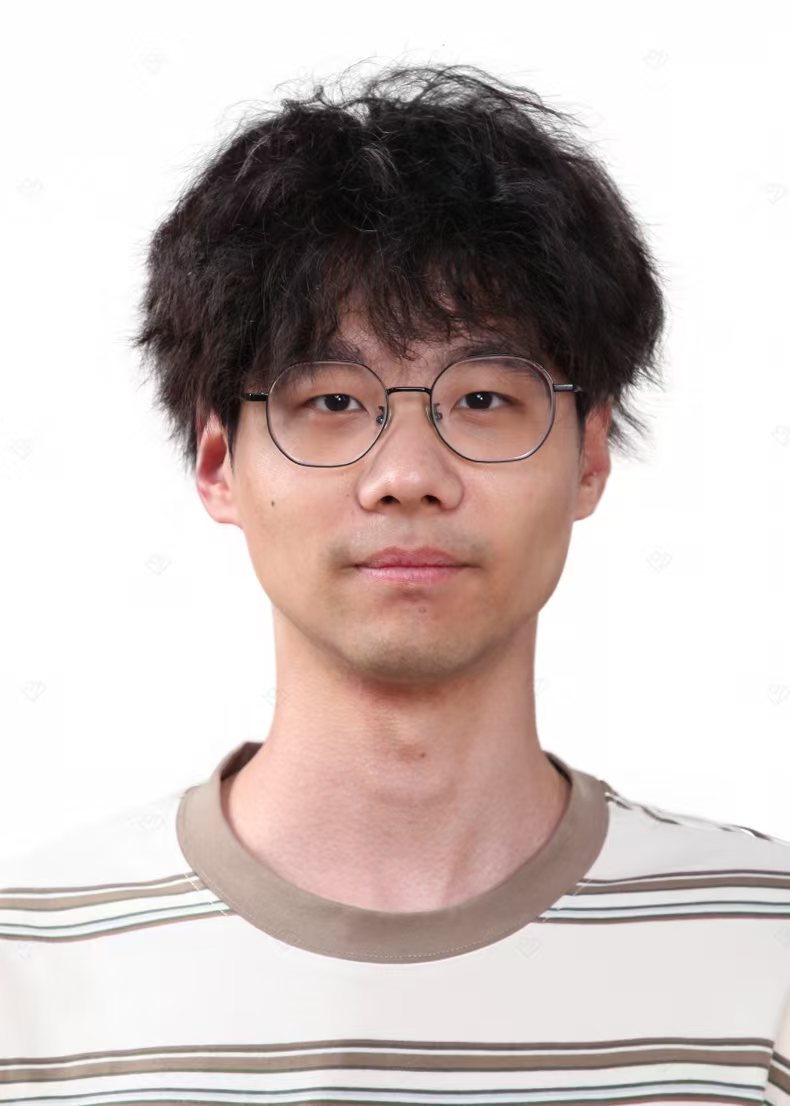}}
\noindent {\bf Bolin Fu}\ 
received the B.S. degree in 2023 from Xidian University, Xi’an, China, where he is currently working toward the master's degree with the School of Artificial Intelligence, Xidian University. His research interests include multi-modal modeling, image processing and machine learning.}
\vspace{1\baselineskip}

\par\noindent 
\parbox[t]{\linewidth}{
\noindent\parpic{\includegraphics[height=1.5in,width=1in,clip,keepaspectratio]{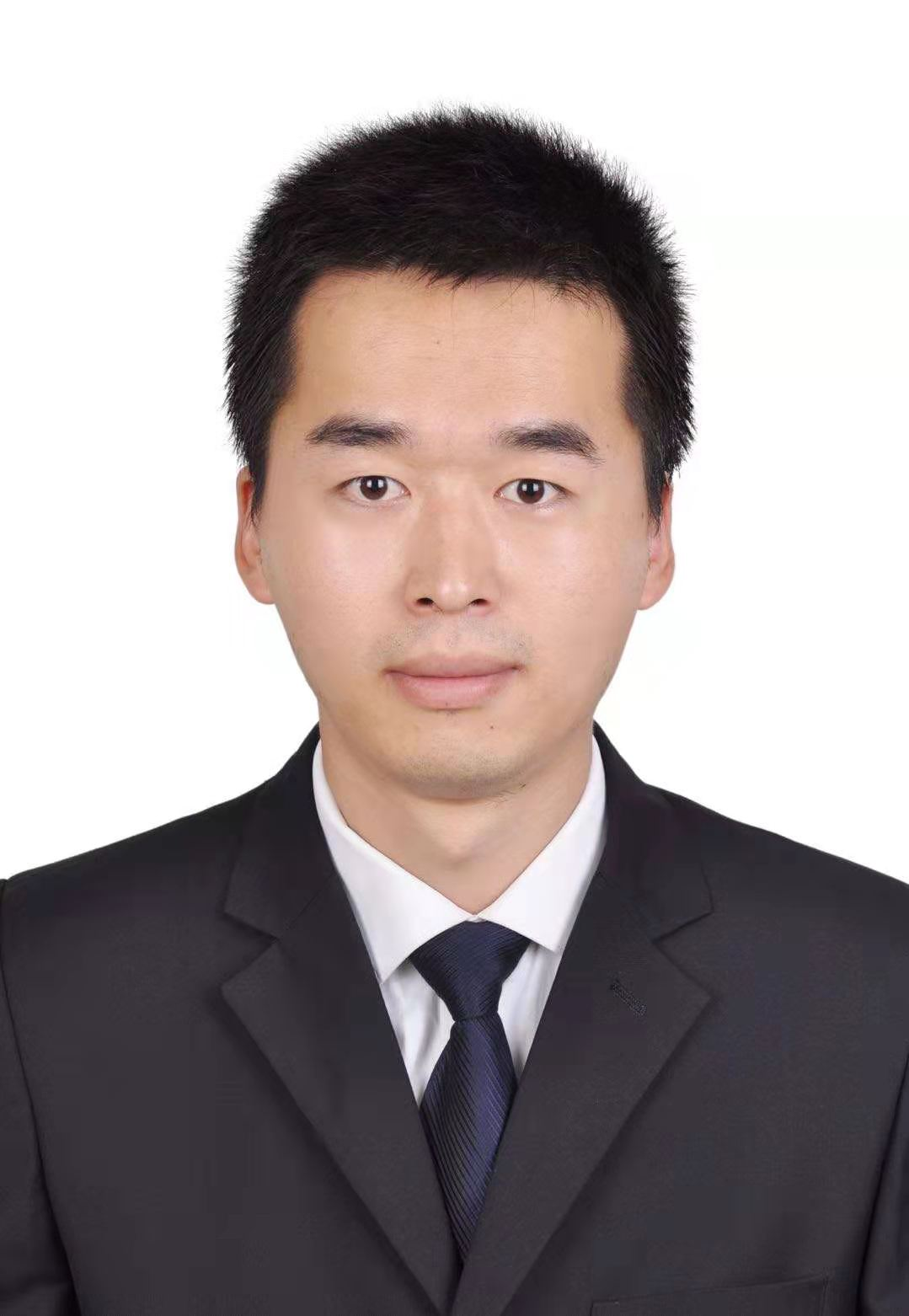}}
\noindent {\bf De Cheng}\
is an associate professor with School
of Telecommunications Engineering, Xidian University, China. He received the B.S. and Ph.D. degrees
from Xi’an Jiaotong University, Xi’an, China, in
2011 and 2017, respectively. From 2015 to 2017, he
was a visiting scholar in Carnegie Mellon University,
Pittsburgh, USA. His research interests include pattern recognition, machine learning, and multimedia
analysis.}
\vspace{1\baselineskip}

\par\noindent 
\parbox[t]{\linewidth}{
\noindent\parpic{\includegraphics[height=1.5in,width=1in,clip,keepaspectratio]{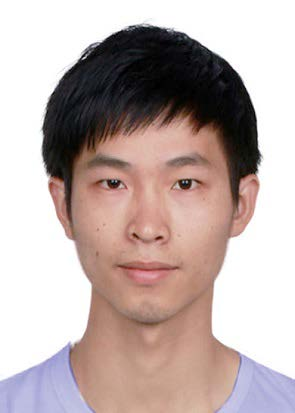}}
\noindent {\bf Lechao Cheng}\
 received the Ph.D. degree from the College of Computer
Science and Technology at Zhejiang University,
Hangzhou, China, in 2019. He is currently an
Associate Professor at the School of Computer
Science and Information Engineering, Hefei University of Technology, Hefei, China. He has
contributed more than 40 research papers to
renowned academic journals and conferences
such as IJCV, TMI, TCyb, TMM, CVPR, AAAI, 
IJCAI, and ACM MM. His research area centers 
around vision knowledge transfer in deep learning.}
\vspace{1\baselineskip}

\par\noindent 
\parbox[t]{\linewidth}{
\noindent\parpic{\includegraphics[height=1.5in,width=1in,clip,keepaspectratio]{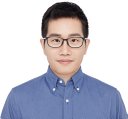}}
\noindent {\bf Guanbin Li}\
received the Ph.D. degree in computer science from the University of Hong Kong, Hong Kong, in 2016. He is currently an Associate Professor with the School of Computer Science and Engineering, Sun Yat-sen University, Guangzhou, China. He has authored and coauthored more than 80 papers in top-tier academic journals and conferences. His research interests include computer vision, image processing, and deep learning. Dr. Li is a recipient of ICCV 2019 Best Paper Nomination Award. He serves as an area chair for the conference of CVPR2024. He has been a reviewer for numerous academic journals and conferences such as TPAMI, IJCV, TIP, TMM, CVPR and PR, ICCV, ECCV, and NIPS.
}
\end{document}